\documentclass{article}

\usepackage[numbers]{natbib}
\usepackage[preprint]{neurips_2023}

\usepackage[utf8]{inputenc} % allow utf-8 input
\usepackage[T1]{fontenc}    % use 8-bit T1 fonts
\usepackage{hyperref}       % hyperlinks
\usepackage{url}            % simple URL typesetting
\usepackage{booktabs}       % professional-quality tables
\usepackage{amsfonts}       % blackboard math symbols
\usepackage{nicefrac}       % compact symbols for 1/2, etc.
\usepackage{microtype}      % microtypography
\usepackage{xcolor}         % colors
\usepackage{array}
\usepackage{bm}
\usepackage{graphicx}
\usepackage{multirow}
\usepackage{enumitem}
\usepackage{subfig}
\usepackage{float}

\newcommand*{\vertbar}{\rule[-1ex]{0.5pt}{2.5ex}}

\title{Representations Matter: Embedding Modes of Large Language Models using Dynamic Mode Decomposition}

\author{%
  Mohamed Akrout \\
  University of Manitoba, AIP Labs\\
  \texttt{akroutm@myumanitoba.ca} \\
}

\begin{document}

\maketitle

\begin{abstract}

Existing large language models (LLMs) are known for generating ``hallucinated'' content, namely a fabricated text of plausibly looking, yet unfounded, facts. To identify when these hallucination scenarios occur, we examine the properties of the generated text in the embedding space. Specifically, we draw inspiration from the dynamic mode decomposition (DMD) tool in analyzing the pattern evolution of text embeddings across sentences. We empirically demonstrate how the spectrum of sentence embeddings over paragraphs is constantly low-rank for the generated text, unlike that of the ground-truth text. Importantly, we find that evaluation cases having LLM hallucinations correspond to ground-truth embedding patterns with a higher number of modes being poorly approximated by the few modes associated with LLM embedding patterns. In analogy to near-field electromagnetic evanescent waves, the embedding DMD eigenmodes of the generated text with hallucinations vanishes quickly across sentences as opposed to those of the ground-truth text. This suggests that the hallucinations result from both the generation techniques and the underlying representation.
\end{abstract}

\section{Introduction}
Language Language Models (LLMs) have emerged as impressive tools capable of generating coherent and contextually relevant answers. These models, trained on vast amounts of data, demonstrated abilities that include understanding language, providing custom and comprehensive answers, and generating creative content. However, LLMs also tend to generate nonsensical or unfaithful content \cite{ji2023survey}. This phenomenon, referred to as \textit{hallucinations}, is one of the crucial challenges that hinder the success and performance of LLMs.

Various research endeavors have begun proposing ways to overcome hallucinations in different text-generation tasks such as summarization, question-answering, and translation \cite{ji2023survey}. A plethora of tools have been introduced to detect hallucinated content. As one example, fact-verification consists of assessing the truthfulness of the LLM output using evidence retrieved from an external knowledge base \cite{fact-checking-survey}. Other approaches have focused on hallucination detection methods for black-box LLMs (i.e., models accessible only through an API) since, in the absence of information about the internal state of the LLM (e.g., embeddings, token probability, or entropy), it is not possible to use uncertainty metrics to detect hallucinations. A self-evaluation approach was proposed where the LLM is asked to answer if its generated answer is true \cite{self-eval}. Alternatively, the LLM can be used to sample different responses, and hallucinations are detected when there is information inconsistency between the different samples \cite{manakul2023selfcheckgpt}. Using this self-checking approach, it was concluded that the hallucination problem is not inherently one of training or representation but is rather one of generation, given that LLMs contain enough information to reduce the hallucination rate.

Here, we call into question this conclusion by observing the embedding properties of the generated text that highlight the importance of the learned representations by LLMs. Our observations do not offer a complete answer about the importance level of representations in generating hallucinations. However, it does aim to balance research discussions and reintroduce questions about representations back to the fold. In this work, we assume that we have access to a black-box LLM and that no external knowledge base is available.

To emphasize the importance of learned representations by LLMs, we resort to the dynamic mode decomposition (DMD) \cite{kutz2016dynamic} method to find the modes of both generated and ground-truth texts (e.g. Wikipedia). While the standard goal of DMD is extracting key \textit{spatiotemporal features} of high-dimensional systems, in this work, we employ DMD to obtain the \textit{embedding properties over sentences}. We do this by substituting the time and system state dimensions with the sentence and embedding dimensions, respectively. It is important to highlight that DMD approximates the modes of the so-called Koopman operator \cite{kutz2016dynamic}. While the Koopman operator is linear and infinite-dimensional, it does not linearize the system dynamics and represents the flow of the dynamical system on measurement functions as an infinite-dimensional operator \cite{rowley2009spectral}. For this reason, the application of DMD in the text embedding space accounts for the non-linear behavior of the embedding dimensions as they evolve through sentences.

\section{Dynamic mode decomposition}
\subsection{Background}
Consider a non-linear dynamical system whose state $\bm{x}_t\in\mathbb{R}^N$ at time $t$ is governed by
\begin{equation}\label{eq:dynamics}
    \frac{\textrm{d}\bm{x}_t}{\textrm{d}t} = \bm{f}\left(\bm{x}_t,t\right),
\end{equation}
where $\bm{f}(\cdot)$ represents the dynamics. For many applications (e.g., fluid dynamics, structural vibrations, neuroscience), the dynamics $\bm{f}$ is difficult to approximate, or simply unknown, and only state measurements $\{\bm{x}_t\}_{t=1}^T$ are available. In this case, the DMD method takes the equation-free perspective by approximating (\ref{eq:dynamics}) by its discrete-time version sampled every $\Delta t$ in time as
\begin{equation}\label{eq:dynamics-approx}
    \bm{x}_{t+1} = \bm{A}\,\bm{x}_{t},
\end{equation}
where $\bm{A}$ refers to the dynamics matrix. The DMD algorithm was devised to collect state measurements from a dynamical system at regularly spaced time intervals by defining the two matrices:

\begin{equation}\label{eq:X-Xp}
\bm{X} = 
\left[
  \begin{array}{cccc}
    \vertbar & \vertbar &        & \vertbar \\
    \bm{x}_1    & \bm{x}_2    & \ldots & \bm{x}_{T-1}    \\
    \vertbar & \vertbar &        & \vertbar 
  \end{array}
\right],~\textrm{and}~~
\bm{X}^\prime = 
\left[
  \begin{array}{cccc}
    \vertbar & \vertbar &        & \vertbar \\
    \bm{x}_2    & \bm{x}_3   & \ldots & \bm{x}_T    \\
    \vertbar & \vertbar &        & \vertbar 
  \end{array}
\right].
\end{equation}

\noindent Then, DMD finds the optimal local linear approximation of the non-linear dynamical system by satisfying the approximation
\begin{equation}\label{eq:approx-DMD}
    \bm{X}^\prime \approx \bm{A}\,\bm{X}.
\end{equation}
Here, the DMD modes (also called dynamic modes) are the eigenvectors of $\bm{A}$, and each DMD mode corresponds to a particular eigenvalue of $\bm{A}$. The DMD eigenvalues provide information about the temporal behavior of the
spatial structure described by the corresponding eigenvectors.

\subsection{From time dynamics to sentence-wise dynamics}

\begin{figure}[!ht]
\centering
\includegraphics[scale=0.47]{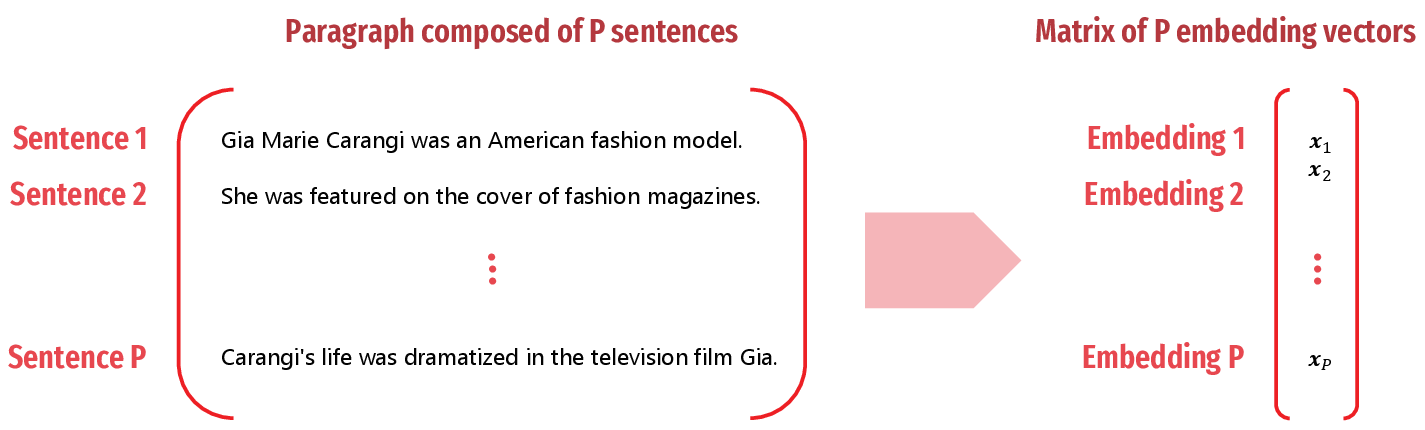}
\caption{The construction of the embedding matrix of a paragraph using a word embedding model.}
\label{fig:conversion}
\vspace{-0.2cm}
\end{figure}

The DMD was originally designed to collect data measurements across different time instants. However, the time dimension in the mathematical derivation of DMD is a dummy dimension, and one can substitute it with another variable depending on the analysis at hand. For NLP use cases, the matrices $\bm{X} \in \mathbb{R}^{N \times P}$ and $\bm{X}^\prime \in \mathbb{R}^{N \times P}$ can be constructed from a paragraph composed of $P$ sentences with each sentence having a text embedding state $\bm{x}_p \in \mathbb{R}^N$ as shown in Figure \ref{fig:conversion}.

\noindent By doing so, the induced DMD eigenmodes and eigenvalues convey information about the \textit{sentence-wise dynamics of the text embedding} instead of the standard \textit{time dynamics of the state measurement}.

\section{Experiments}
We evaluate the text embedding properties of both generated text and ground-truth text using GPT-3 (text-davinci-003) on individuals from the Wikibio dataset \cite{lebret2016neural}. Specifically, we use a sub-dataset of Wikibio from \cite{manakul2023selfcheckgpt} containing 238 articles from the top 20\% of articles by length. This importantly ensures the mitigation of more obscure concepts being incorporated. It is a sub-dataset of the first paragraph (along with tabular information) of Wikipedia biographies\footnote{\href{https://huggingface.co/datasets/potsawee/wiki_bio_gpt3_hallucination}{\texttt{https://huggingface.co/datasets/potsawee/wiki\_bio\_gpt3\_hallucination}}}, and the GPT-3 output was generated in \cite{manakul2023selfcheckgpt} using the prompt ``\texttt{This is a Wikipedia passage about \{concept\}}''. Moreover, each sentence of each data sample generated by GPT-3 was classified into one of three annotation categories:
\begin{itemize}[leftmargin=*]
    \item \textit{major inaccurate} when the sentence is unrelated to the topic and hence is a hallucination sample,
    \item \textit{minor inaccurate} when the sentence is related to the topic but contains non-factual information,
    \item \textit{accurate} when the sentence describes factual information.
\end{itemize}

\subsection{Data processing}\label{eq:data-processing}
For each data sample, we map both the ground-truth Wiki and the GPT-3 generated paragraphs to a 384-dimensional embedding vector space using the HuggingFace sentence embeddings API of the sentence-transformers model \texttt{all-MiniLM-L6-v2}\footnote{\href{https://huggingface.co/sentence-transformers/all-MiniLM-L6-v2}{\texttt{https://huggingface.co/sentence-transformers/all-MiniLM-L6-v2}}}. To obtain a single label for each paragraph, if all sentence annotations agree, we use the paragraph label. However, if annotations differ from one sentence to another, we use the most frequent annotation as the paragraph label. If multiple annotations have the same occurrence frequency, we use the worse-case label as the paragraph label. For example, the sentence annotation list \{minor inaccurate, minor inaccurate, major inaccurate, major inaccurate\} is mapped to the paragraph annotation \{major inaccurate\}.

\begin{figure}[!h]
\centering
\includegraphics[scale=0.28]{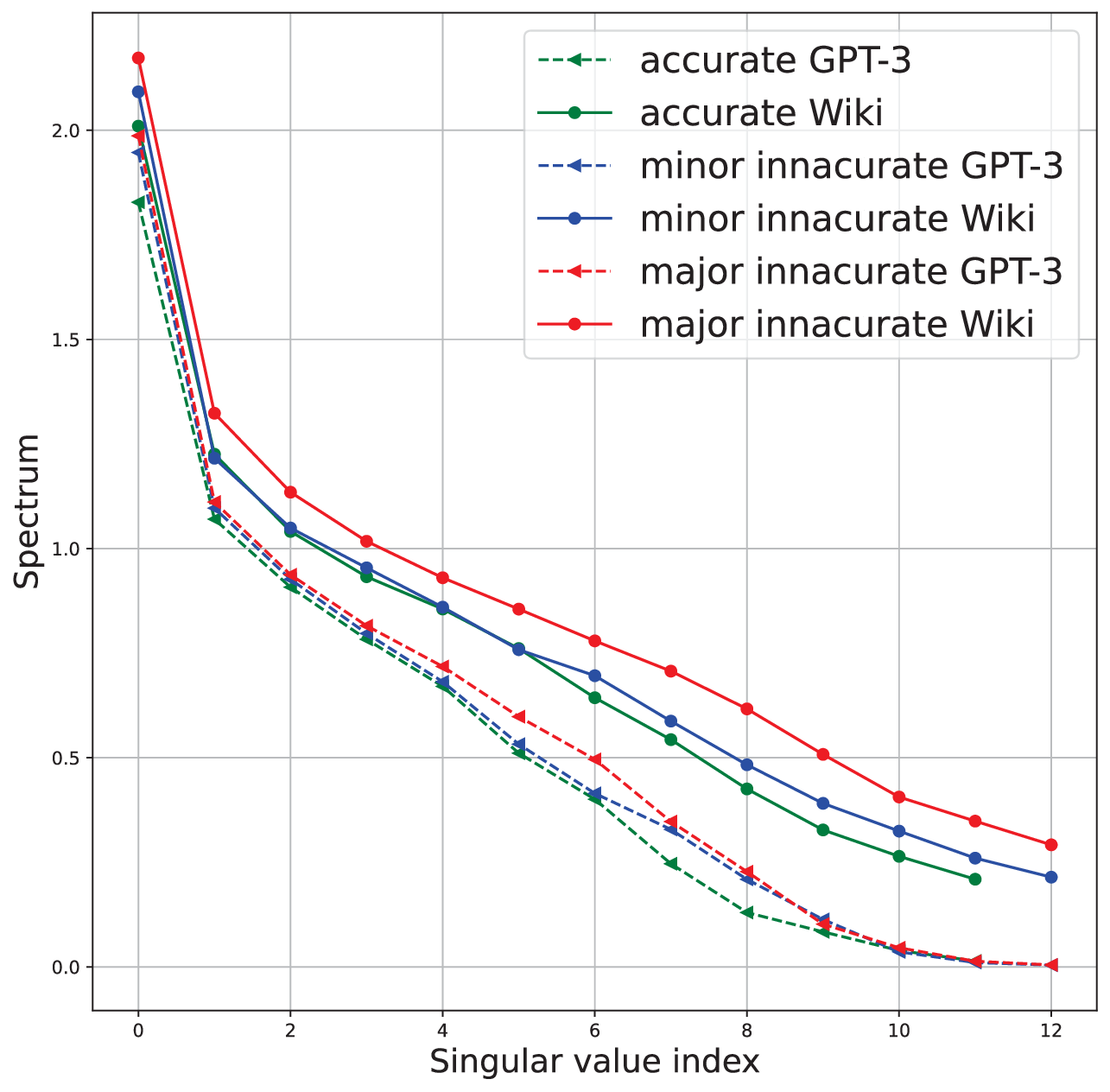}
\caption{The average spectrum of the Wiki and GPT-3 text embedding matrices $\bm{X}$ associated with major inaccurate, minor inaccurate, and accurate data samples.}
\label{fig:embedding-rank}
\end{figure}

We start by stacking all the sentences' embeddings of each paragraph into an embedding matrix. Then, we plot in Figure \ref{fig:embedding-rank} the average spectrum of data samples associated with each of the three annotation categories. There, it is seen that the rank of the Wiki embedding is higher than that of the GPT-3 generated embedding. This means that the number of linearly independent embedding vectors over sentences is higher for Wiki text than for GPT-3 generated text.

\noindent Moreover, it is interesting to note that the  higher hallucination of GPT-3 (e.g., major inaccurately generated samples) is associated with a higher rank of the ground-truth Wiki embeddings. This is opposed to the rank of the GPT-3 embeddings which has a smaller rank for hallucinated and partially hallucinated, or non-hallucinated samples.

\subsection{DMD analysis}
We use the \texttt{PyDMD} package \cite{demo2018pydmd} to examine the modes and dynamics of the text embeddings. We also use the optimal rank method in \cite{gavish2014optimal} to approximate the best-fit matrix $\bm{A}$ in (\ref{eq:approx-DMD}).\footnote{This can be done by letting the parameter $\texttt{svd\_rank=0}$ in the constructor of the $\texttt{DMD}$ class.}.

\begin{figure}[h!]
    \hspace{-0.2cm}
    \subfloat[Wiki, major inaccurate.]{{
    \includegraphics[scale=0.205]{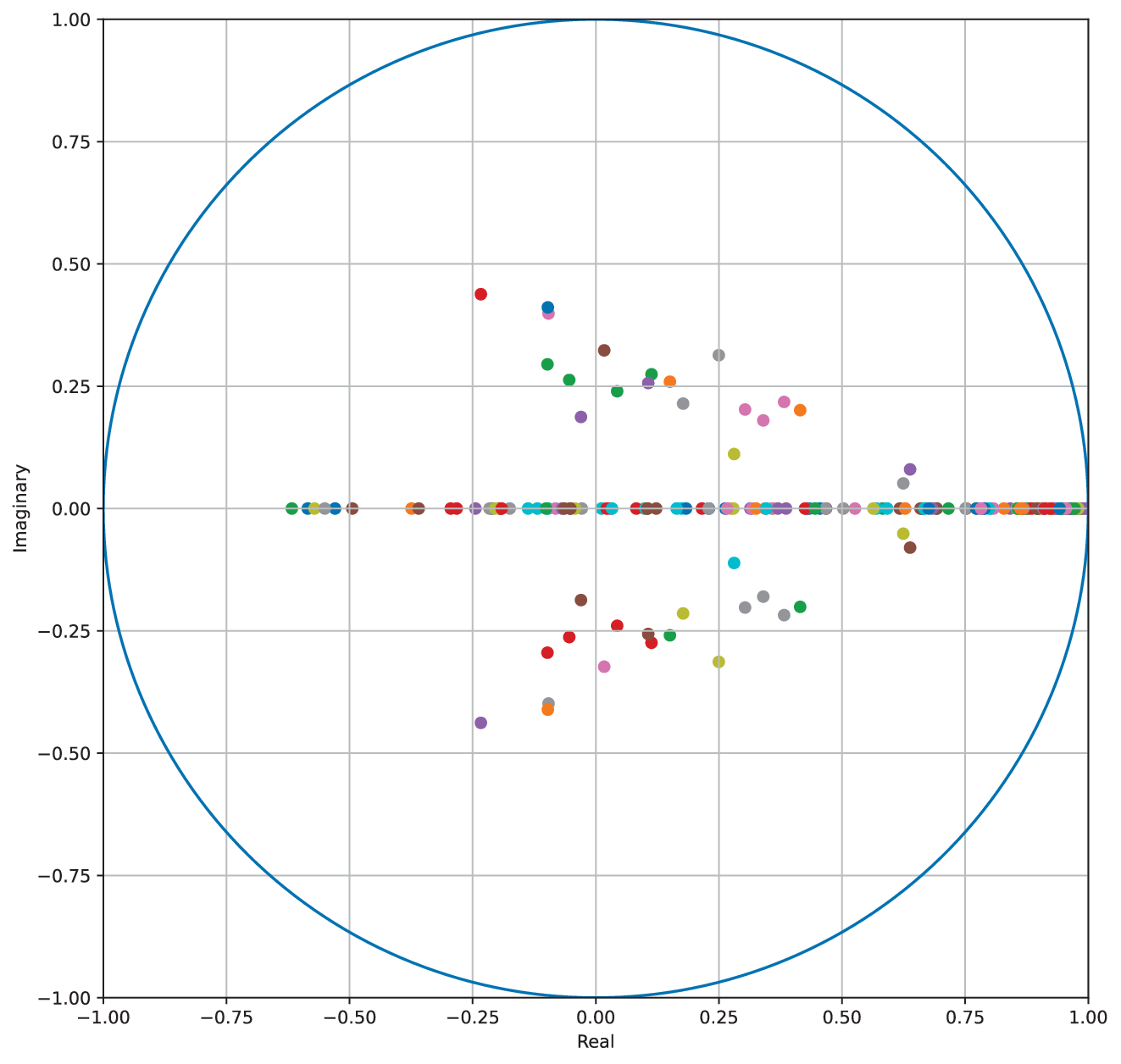}
    \label{fig:major-eigen-wiki}}}%
    \subfloat[Wiki, minor inaccurate.]{{
    \includegraphics[scale=0.205]{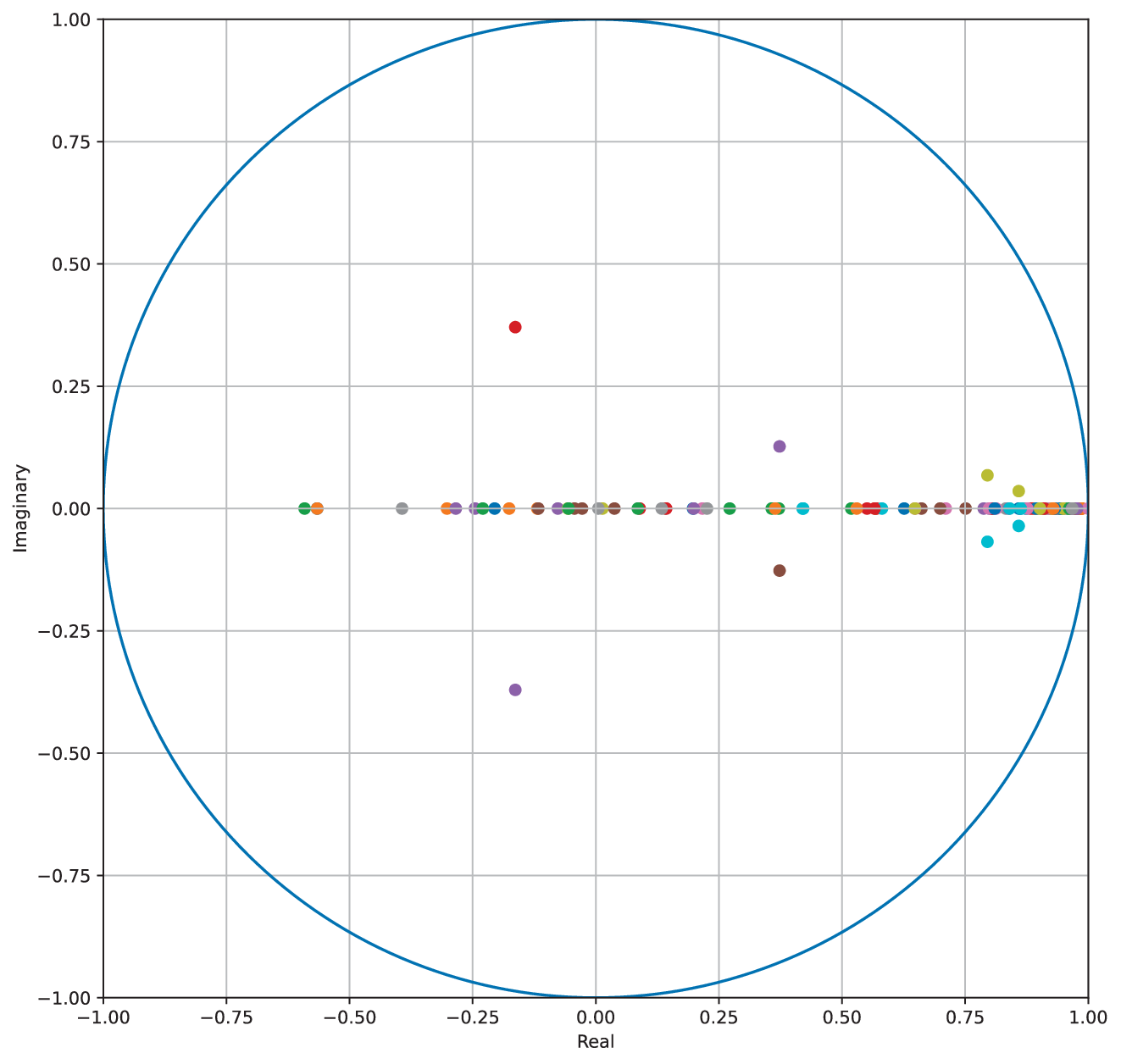}
    \label{fig:minor-eigen-wiki}}}%
    \subfloat[Wiki, accurate.]{{
    \includegraphics[scale=0.205]{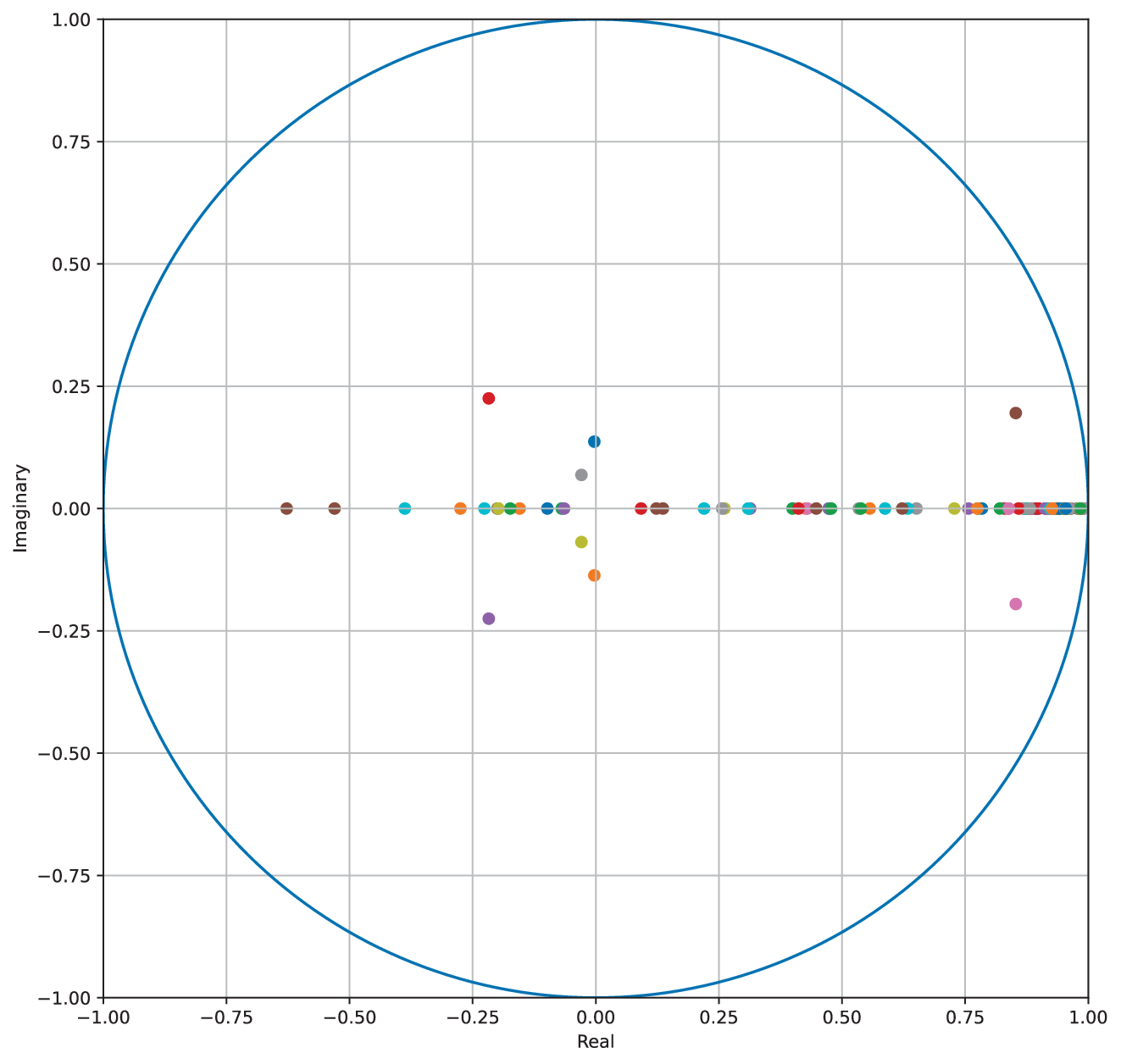}
    \label{fig:acc-eigen_wiki}}}
    \caption{DMD eigenmodes of the text embedding of the wiki text for (a) major inaccurate samples, (b) minor inaccurate samples, and (c) accurate samples.}
    \label{fig:eigen-dmd-wiki}
    \vspace{-0.2cm}
\end{figure}

\begin{figure}[h!]
    \hspace{-0.2cm}
    \subfloat[GPT-3, major inaccurate.]{{
    \includegraphics[scale=0.205]{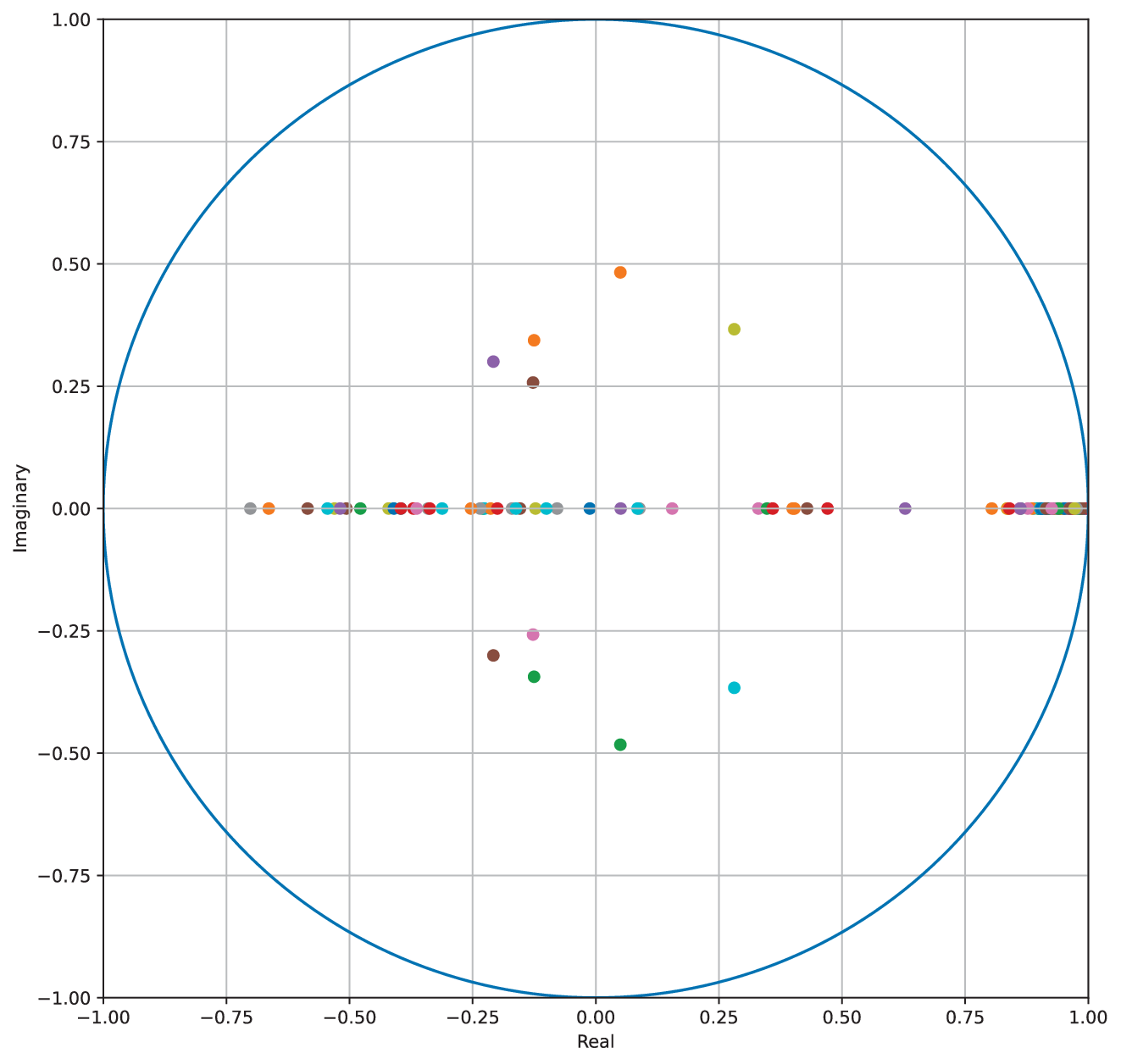} 
    \label{fig:major-eigen-gpt3}}}%
    \subfloat[GPT-3, minor inaccurate.]{{
    \includegraphics[scale=0.205]{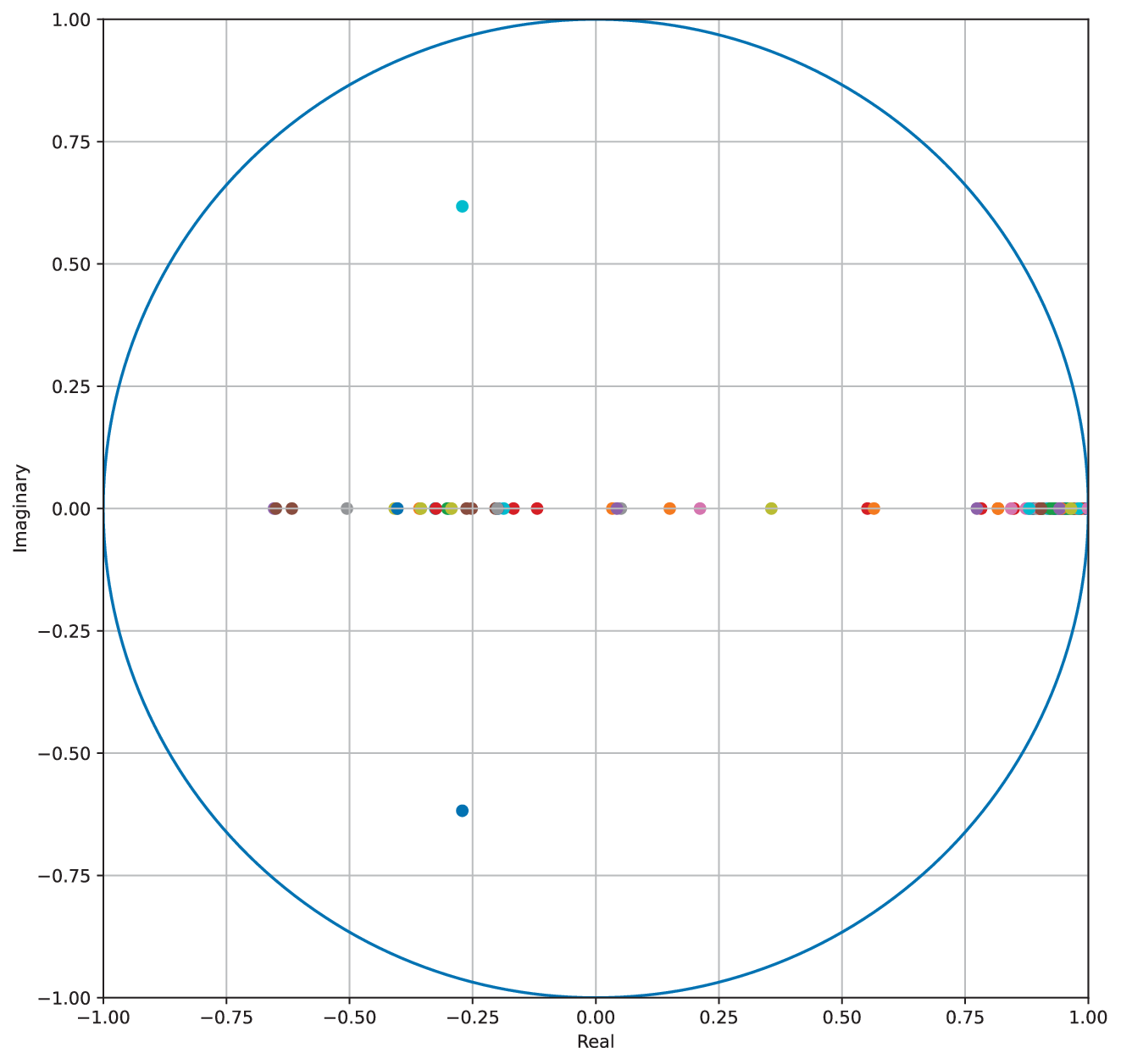}
    \label{fig:minor-eigen-gpt3}}}%
    \subfloat[GPT-3, accurate.]{{
    \includegraphics[scale=0.205]{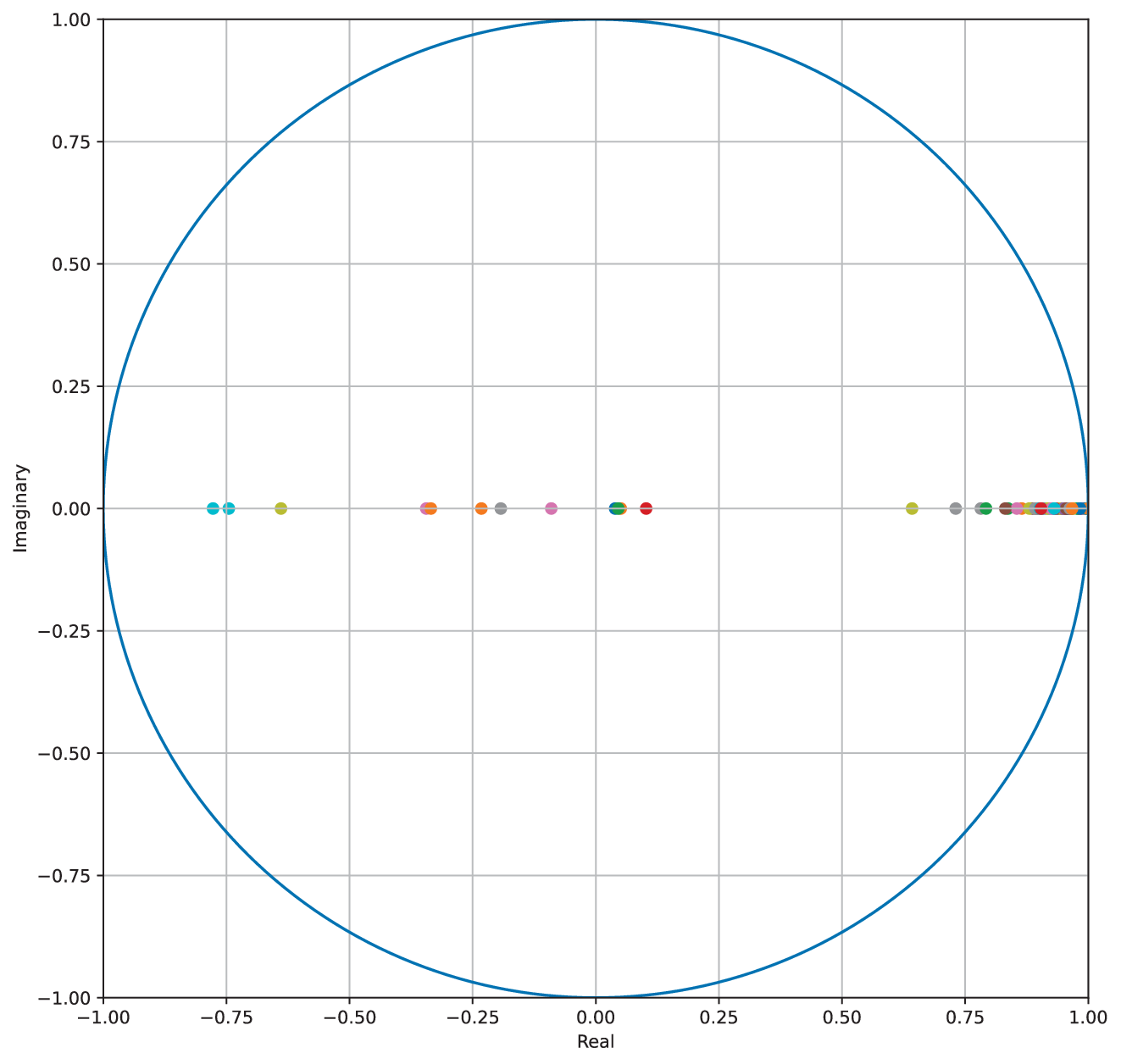}
    \label{fig:acc-eigen-gpt3}}}
    \caption{DMD eigenmodes of the text embedding of the GPT-3 generated text for (a) major inaccurate samples, (b) minor inaccurate samples, and (c) accurate samples.}
    \label{fig:eigen-dmd-gpt3}
\end{figure}

Figures \ref{fig:eigen-dmd-wiki} and \ref{fig:eigen-dmd-gpt3} depict the DMD eigenvalues of the modes associated with Wiki and GPT-3 text, respectively, for paragraphs with the three annotations described in Section \ref{eq:data-processing}. First, it is observed that all the eigenvalues in the complex plane are inside the unit circle. This suggests that the embedding eigenmodes describe transient embedding responses decreasing across sentences.

We also note how a significant number of the eigenvalues of the Wiki text associated with hallucinated samples in Figure \ref{fig:major-eigen-wiki} are complex. This is to be opposed to the eigenvalues of $i)$ Wiki text having minor inaccuracies or being accurate (cf. Figures \ref{fig:minor-eigen-wiki} and \ref{fig:acc-eigen_wiki}), and $ii)$ those of the generated GPT-3 text in \ref{fig:eigen-dmd-gpt3} independently of its accuracy. 

Unlike the discrete Fourier transform (DFT) whose eigenvalues are roots of unity, the DMD eigenvalues are characterized by both a frequency and a magnitude, which together correspond to a complex frequency. The latter characterizes important physical properties (e.g., periodicity, oscillation, damping) in real-world dynamical systems \cite{krake2021visualization}. In loose analogical reference to electromagnetics (EM), one would compare embedding modes vanishing over sentences to EM waves radiated in space. The embedding modes associated with complex eigenvalues correspond to far-field propagation modes, while the real-valued eigenvalues are associated with near-field ones (e.g., evanescent waves) which vanish rapidly in space.

By contrasting Figures \ref{fig:major-eigen-wiki} and \ref{fig:major-eigen-gpt3}, it is seen that generated samples with hallucinations correspond to generated embeddings with a few modes propagating across sentences as opposed to the Wiki embeddings, where many more embeddings survive across sentences. In other words, hallucinations occur when a few embedding modes generated by GPT-3 are approximating a larger set of embedding modes of the Wiki text. Confirmingly, little to no hallucinations are observed when the Wiki text's embeddings have only a few modes. In such scenarios, this suggests that the ``few mode embedding nature'' of the GPT-3 generation does not hallucinate and approximates well the embedding modes of the ground truth. Reverting back to the prior EM analogy, approximating the higher numbers of modes in the near-field with the lower numbers of modes in the far-field is called ``electromagnetic inversion'', which looks conceptually similar to the phenomenon of hallucinations in the embedding space. We summarize the analogy between the hallucination with LLMs and the EM wave theory in Table \ref{tab:EM-analogy}.

\begin{table}[ht]
\scriptsize{
\caption{The analogy between the electromagnetic modes and the DMD text embedding modes for scenarios with or without hallucination.}
\hspace{2cm}
\begin{tabular}[t]{ccc}
\cmidrule{2-3}
& &\hspace{-4cm}\textbf{Terminology}\\
\cmidrule{2-3}
& \textbf{Electromagnetics}&\textbf{NLP}\\
\midrule
\multirow{4}{*}{\vspace{-0.45cm}\textit{hallucination}}& radiation source&fact\\ \cmidrule{2-3}
&evanescent modes&DMD modes of the ground-truth text embedding\\ \cmidrule{2-3}
&far-field modes& DMD modes of the LLM text embedding\\
\cmidrule{2-3}
&inversion & hallucination\\\midrule\midrule
\multirow{2}{*}{\vspace{-0.15cm} \textit{no hallucination}}& radiation source&fact\\ \cmidrule{2-3}
&far-field modes& DMD modes of LLM/ground-truth text embedding\\
\bottomrule
\end{tabular}
\label{tab:EM-analogy}
}
\end{table}

\begin{figure}[h!]
    \hspace{-0.2cm}
    \subfloat[Wiki, major inaccurate.]{{
    \includegraphics[scale=0.208]{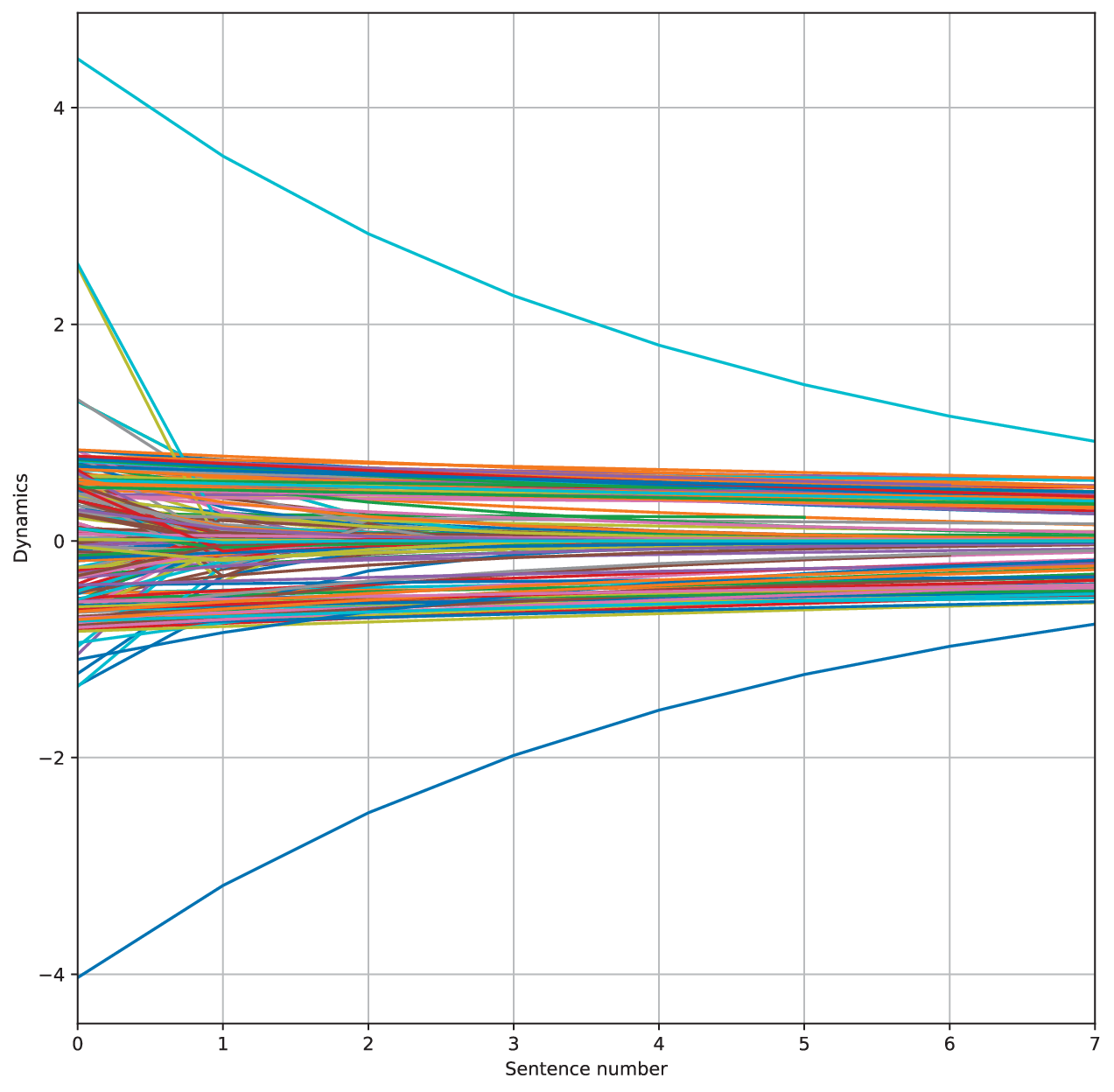} 
    \label{fig:major-dynamics-wiki}}}%
    \subfloat[Wiki, minor inaccurate.]{{
    \includegraphics[scale=0.208]{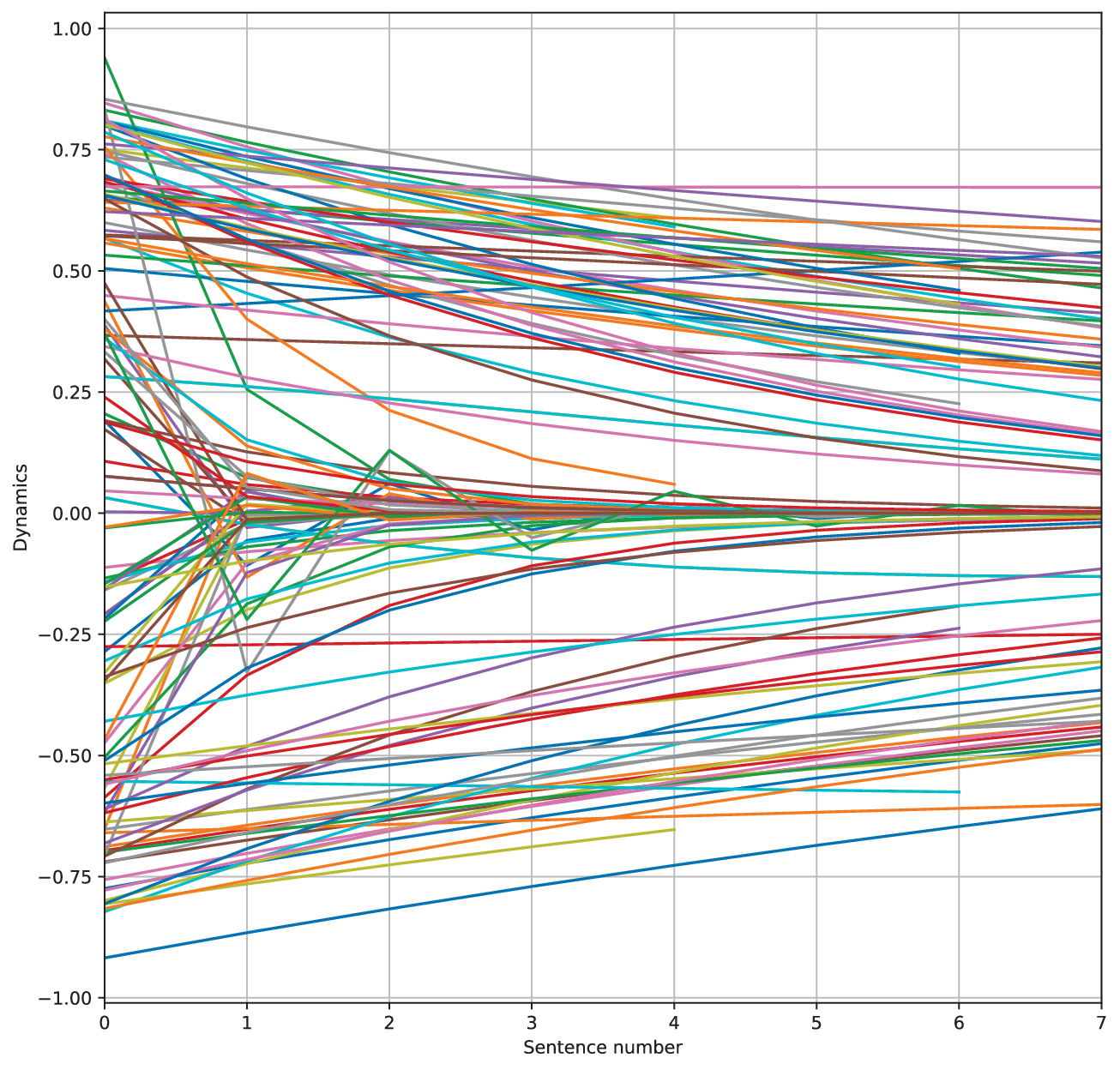}
    \label{fig:minor-dynamics-wiki}}}%
    \subfloat[Wiki, accurate.]{{
    \includegraphics[scale=0.208]{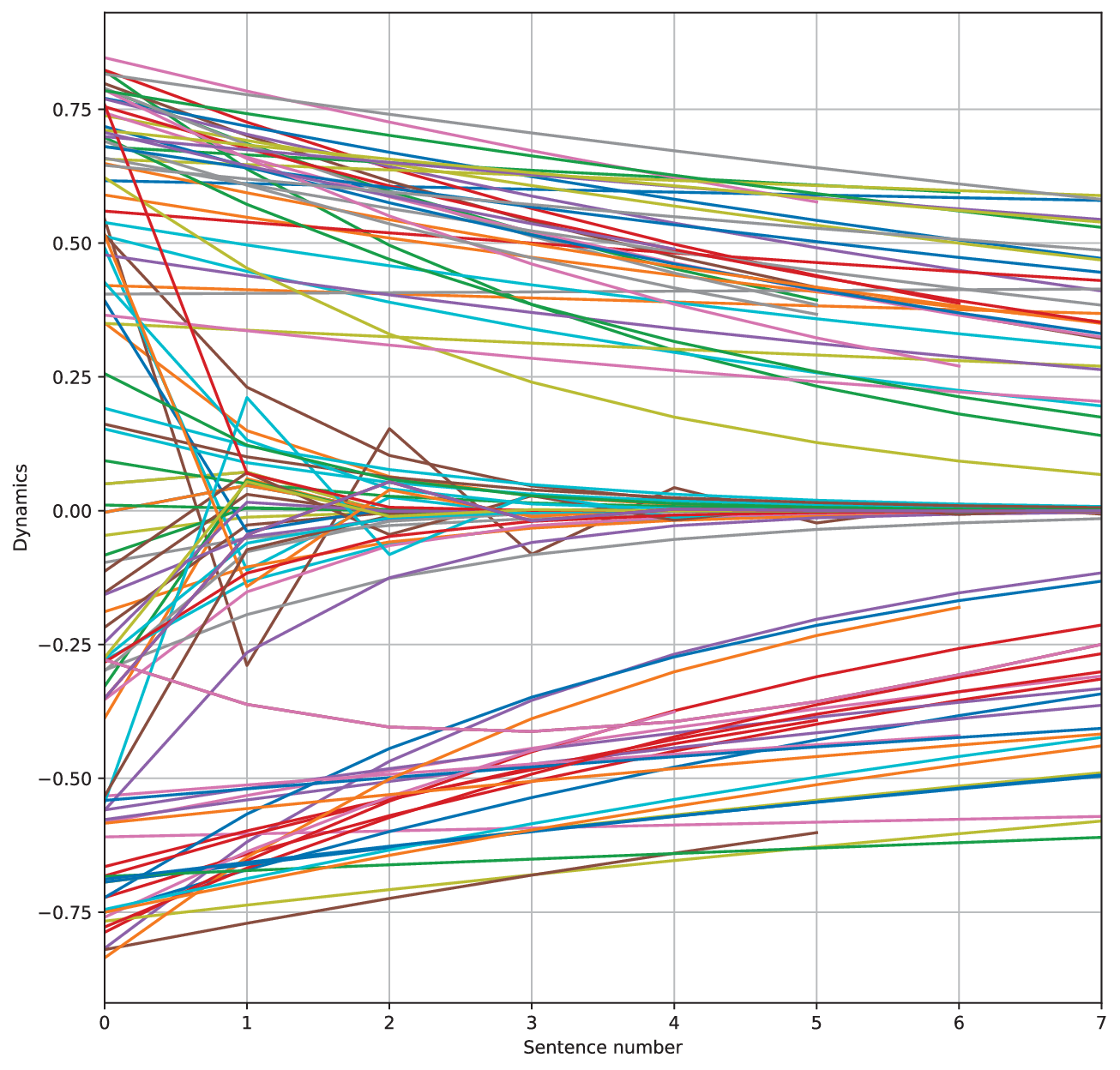}
    \label{fig:acc-dynamics-wiki}}}
    \caption{DMD mode dynamics of the text embedding of the wiki text for (a) major inaccurate samples, (b) minor inaccurate samples, and (c) accurate samples.}
    \label{fig:dynamics-dmd-wiki}
\end{figure}

\begin{figure}[h!]
    \hspace{-0.2cm}
    \subfloat[GPT-3, major inaccurate.]{{
    \includegraphics[scale=0.208]{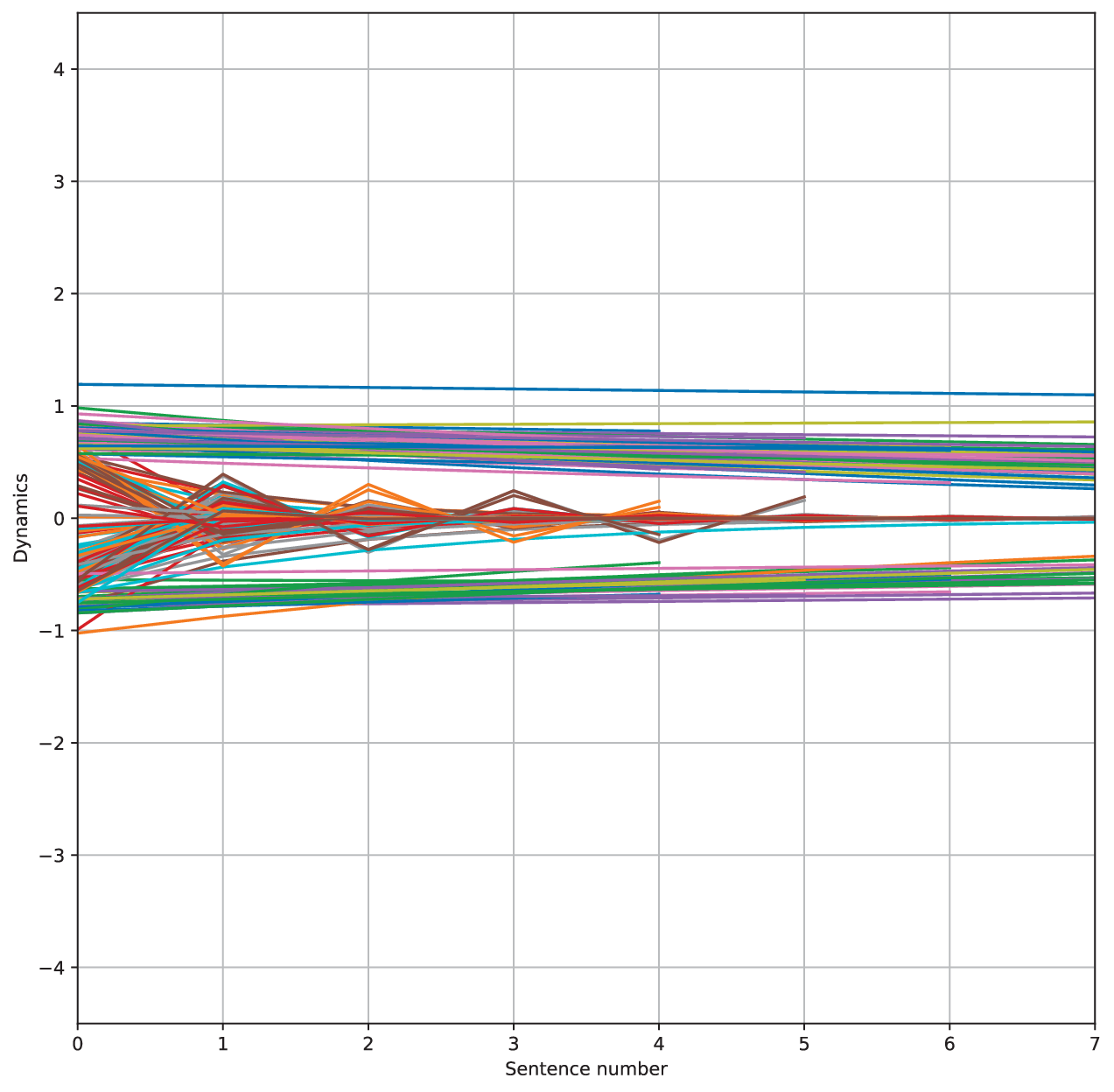} 
    \label{fig:major-dynamics-gpt3}}}%
    \subfloat[GPT-3, minor inaccurate.]{{
    \includegraphics[scale=0.208]{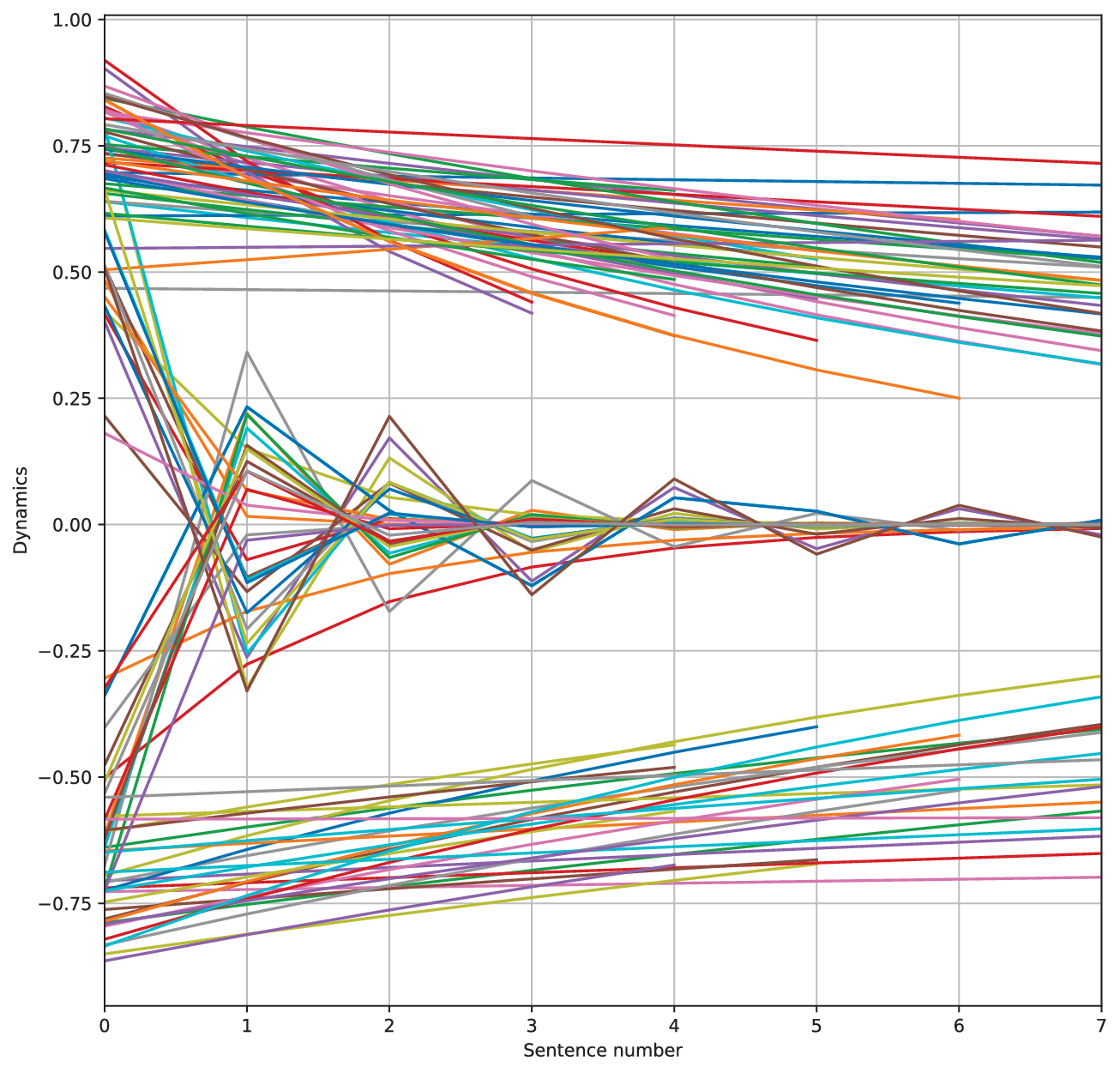}
    \label{fig:minor-dynamics-gpt3}}}%
    \subfloat[GPT-3, accurate.]{{
    \includegraphics[scale=0.208]{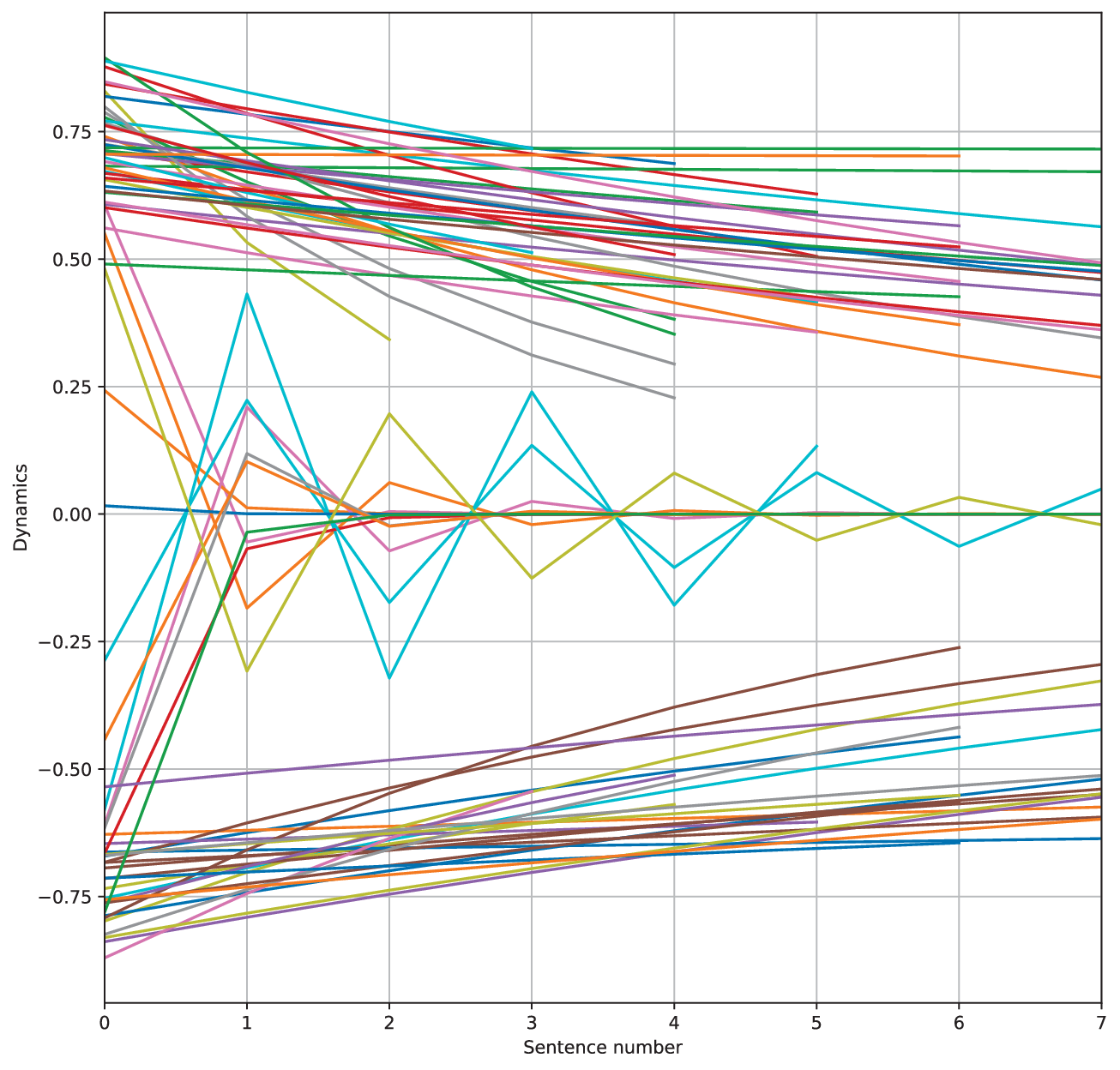}
    \label{fig:acc-dynamics-gpt3}}}
    \caption{DMD mode dynamics of the text embedding of the GPT-3 generated text for (a) major inaccurate samples, (b) minor inaccurate samples, and (c) accurate samples.}
    \label{fig:dynamics-dmd-gpt3}
    \vspace{-0.2cm}
\end{figure}

Finally, we plot the dynamics of all the embedding modes for both Wiki and GPT-3 generated text in Figures \ref{fig:dynamics-dmd-wiki} and \ref{fig:dynamics-dmd-gpt3}.

\noindent There, we observe the decreasing behavior of the modes for accurate and minor inaccurate samples, confirming why all the eigenvalues were inside the unit circle. For major inaccurate samples, some modes' dynamics are consistently present across sentences as they slowly decrease, and appear to be constant. It is worth noting how the number of these gradually decreasing modes is much higher for the embeddings of the Wiki test. This confirms how the fewer non-vanishing embedding modes of the GPT-3 text can be a good approximation of those of the Wiki text only when the latter itself has a few modes as well.

\section{Conclusion \& future work}

This work sought to find a connection in the embedding space between the dynamic mode decomposition modes of the GPT-3 generated text and those of its ground-truth Wiki text. We empirically found that hallucination occurs when the embedding modes of the Wiki text have a significant number of modes that cannot be well approximated by those of the GPT-3 text. One limitation of this work is that it does not revisit the prompting of the annotated dataset of Wikibio from \cite{manakul2023selfcheckgpt}. Questions about the prompting strategies are important to answer in future studies as they uncover new properties of the text embeddings. Further tests can be conducted on other datasets after annotating them to confirm those findings on multiple contexts, prompts, and generated text.
\small{
\bibliographystyle{unsrt}
\bibliography{refs}
}
\end{document}